\documentclass[10pt,letterpaper]{article}

\usepackage[pagenumbers]{cvpr} 
\usepackage{algorithm2e}
\usepackage{amsmath}
\usepackage{arydshln}
\usepackage{bm}
\usepackage{booktabs}
\usepackage{color}
\usepackage{dashrule}
\usepackage{dsfont}
\usepackage{graphicx}
\usepackage{makecell}
\usepackage{multirow}
\usepackage{verbatim}
\usepackage{array} 
\usepackage{color}

\usepackage{amssymb}
\newcommand{\cmark}{\checkmark}
\usepackage{newfloat}
\usepackage{listings}
\definecolor{cvprblue}{rgb}{0.21,0.49,0.74}
\usepackage[pagebackref,breaklinks,colorlinks,allcolors=cvprblue]{hyperref}

\usepackage{color}
\newcommand{\fzl}[1]{{\color{black}#1}}

\usepackage{color}

\title{SEW: Self-calibration Enhanced Whole Slide Pathology Image Analysis}

\author{Haoming Luo\\
Department of Software\\
Zhejiang University\\
{\tt\small haoming\_luo@zju.edu.cn}
\and
Xiaotian Yu\\
Department of Computer Science\\
and Technology\\
Zhejiang University\\
{\tt\small xiaotian\_yu@zju.edu.cn}
\and
Shengxuming Zhang\\
Department of Computer
\\Science and Technology\\
Zhejiang University\\
{\tt\small zsxm1998@zju.edu.cn}
\and
Jiabin Xia\\
Department of Software\\
Zhejiang University\\
{\tt\small jiabin\_xia@zju.edu.cn}
\and
Jian Yang\\
Department of Software\\
Zhejiang University\\
{\tt\small yang\_jian@zju.edu.cn}
\and
Yuning Sun\\
Department of Software\\
Zhejiang University\\
{\tt\small yuning\_sun@zju.edu.cn}
\and
Jing Zhang\\
The First Affiliated Hospital\\
of Zhejiang University\\
Zhejiang University\\
{\tt\small jzhang1989@zju.edu.cn}
\and
Mingli Song\\
Department of Computer Science
\\and Technology\\
Zhejiang University\\
{\tt\small songml@zju.edu.cn@zju.edu.cn}
\and
Xiuming Zhang\\
The First Affiliated Hospital\\
of Zhejiang University\\
Zhejiang University\\
{\tt\small zhangxiuming@zju.edu.cn}
\and
ZunLei Feng\\
Department of Computer Science
\\and Technology\\
Zhejiang University\\
{\tt\small zunleifeng@zju.edu.cn}
}

\begin{document}
\maketitle
\begin{abstract}
Pathology images are considered the ``gold standard" for cancer diagnosis and treatment, with gigapixel images providing extensive tissue and cellular information. 
Existing methods fail to simultaneously extract global structural and local detail features for comprehensive pathology image analysis efficiently.
To address these limitations, we propose a self-calibration enhanced framework for whole slide pathology image analysis, comprising three components: a global branch, a focus predictor, and a detailed branch. The global branch initially classifies using the pathological thumbnail, while the focus predictor identifies relevant regions for classification based on the last layer features of the global branch. The detailed extraction branch then assesses whether the magnified regions correspond to the lesion area.
Finally, a feature consistency constraint between the global and detail branches ensures that the global branch focuses on the appropriate region and extracts sufficient discriminative features for final identification.
These focused discriminative features prove invaluable for uncovering novel prognostic tumor markers from the perspective of feature cluster uniqueness and tissue spatial  distribution.
Extensive experiment results demonstrate that the proposed framework can \textbf{rapidly} deliver \textbf{accurate} and \textbf{explainable} results for pathological grading and prognosis tasks.
\end{abstract}    
\section{Introduction}
\label{sec:intro}
Pathology images are regarded as the ``gold standard" for cancer diagnosis and treatment due to their rich microscopic cellular and tissue characteristics. These images exhibit a super-large size and a wealth of features, necessitating that pathologists frequently zoom in and out to examine both global structural elements and localized details for accurate diagnosis. However, the extensive size and complexity of these images can result in time-consuming evaluations and may lead to misdiagnoses or missed diagnoses due to the inability to scrutinize every detail.

\fzl{
Some researchers~\cite{DSNet,weaklywang2018} analyzed pathological thumbnails due to the challenges existing deep learning models face in processing large images. Others~\cite{zoommilthandiackal2022differentiable,shao2021transmil} employed Multi-Instance Learning with randomly selected patches for whole slide image analysis; however, the former often lacks detailed tissue and cellular features, while the latter fails to capture global structural information.
}


Consequently, multiple-layer pyramid features are employed for pathology image analysis~\cite{chen2021diagnose,hipt2022scaling}. However, these pyramid features increase the computational time required for processing gigapixel pathology images. 
Furthermore, extensive mixed features often contain a limited number of task-related features alongside a significant proportion of irrelevant features, which can impair the identification performance of the final model.

In this paper, we propose a self-calibration enhanced whole slide pathology image analysis framework, termed SEW, which integrates global features and several critical local features for fast and accurate pathology image analysis.
SEW first classifies images using the global structural features derived from pathology thumbnails. A focus predictor is then employed to identify suspected lesion areas with high probability. Subsequently, these areas are enlarged to extract local detail features, determining whether they correspond to actual lesions. Finally, a feature consistency constraint between the global and local branches is introduced to enhance the global branch’s ability to extract more distinctive features.

The integration of global structural features with local detail features from critical areas effectively mitigates the influence of irrelevant features, thereby improving model accuracy and inference speed. Moreover, unlike most existing methods that analyze image patches, SEW utilizes a superpixel technique to identify initial areas with similar cells and tissues. This approach facilitates feature aggregation and reduces irrelevant feature fusion. Additionally, a pathological prototype vocabulary is constructed using clustered area features, which serves to enforce feature consistency across diverse WSI samples.


\fzl{
With these simplified and discriminative features from the focused regions, the k-means algorithm is employed to uncover distinct clusters of favorable or unfavorable prognosis samples. These unique clusters unveil novel prognostic tumor markers, which are subsequently validated by a pathologist. Moreover, the spatial distribution of distinct tissues further reveals prognostic tumor markers within the two-dimensional spatial realm of the image. 
 }

Our contributions are summarized as follows:
\fzl{We present a self-calibration-enhanced framework for whole-slide pathology image analysis, seamlessly integrating global structural features with pivotal local features to ensure both precision and efficiency. The focus predictor, coupled with a feature consistency constraint strategy, augments the global branch's capacity to extract more distinctive and accurate features. Additionally, we introduce a pathological prototype vocabulary to reinforce feature consistency across diverse WSI samples. Extensive experiments validate that our method achieves state-of-the-art performance in both inference speed and accuracy. 
More importantly, the learned simplify and critical features can effectively prompt new tumor mark finding}.

\section{Related Work}
\label{sec:formatting}

\subsection{Whole Slide Image Analysis}
Existing Whole Slide Image (WSI) analysis methods can be divided into three main categories: thumbnail-based, Multi-Instance Learning (MIL), and pyramid feature-based approaches. Initially, researchers \cite{DSNet,weaklywang2018} utilized pathological thumbnails for WSI analysis to address the challenges of handling extremely large images. However, these methods often result in suboptimal classification due to their inability to capture fine-grained tissue and cellular details.

To capture local detailed features, researchers have employed MIL approaches for WSI analysis \cite{milchikontwe2020multiple,milzhao2020predicting}. 
Specifically, Tellez et al. \cite{tellez2019neural} aggregated features from segmented patches to represent the entire WSI. 
Chen et al. \cite{hipt2022scaling} utilized hierarchical transformers to integrate features across scales. 
Li et al. \cite{wikgli2024dynamic} enhanced structural feature expression by dynamically constructing inter-patch edges. 
Despite these advancements, MIL methods still struggle to capture global structural features.


To integrate global and local features, many researchers adopted pyramid feature for pathology image analysis.
For example, Chen et al.~\cite{hipt2022scaling}  aggregated visual tokens at cell, patch, and region levels in a bottom-up manner to construct slide representations.
Xiang et al. ~\cite{DSNet} used a Dual-Stream Network to obtain representations of multi-scale thumbnail images.
Yu et al.~\cite{smt2024} used a self-reform multilayer transformer to address the time-consuming and space-consuming problem in pathological image analysis. 
Chen et al.~\cite{chen2021diagnose} adopted a tree-based self-supervision to enhance representation learning and suppress contributions of potentially irrelevant patches.
\fzl{
However, these pyramid features elevate the computational time required for processing WSIs, while the extensive mixture of features complicates the learning of critical features for the final task.
}


\subsection{Acceleration of Pathology Image Analysis}
To enhance the training and inference efficiency for pathology images, several researchers have focused on identifying Regions of Interest (ROIs) within WSI for effective analysis.
Lu et al.~\cite{clamlu2021data} employed attention mechanisms to locate ROIs for ultimate classification, using whole-slide labels as supervisory guidance.
Shao et al.~\cite{shao2021transmil} introduced TransMIL, which explores both morphological and spatial information for weakly supervised WSI classification.
Tang et al. ~\cite{tangquadtree} presented the QuadTree method, which deconstructs histopathology images by identifying clinically relevant regions while disregarding less pertinent areas such as empty spaces or connective tissue.
Furthermore, ZoomMIL~\cite{zoommilthandiackal2022differentiable} enables the model to identify informative patches, thereby greatly enhancing inference speed. 

\begin{figure*}[!t]
    \includegraphics[width=\textwidth]{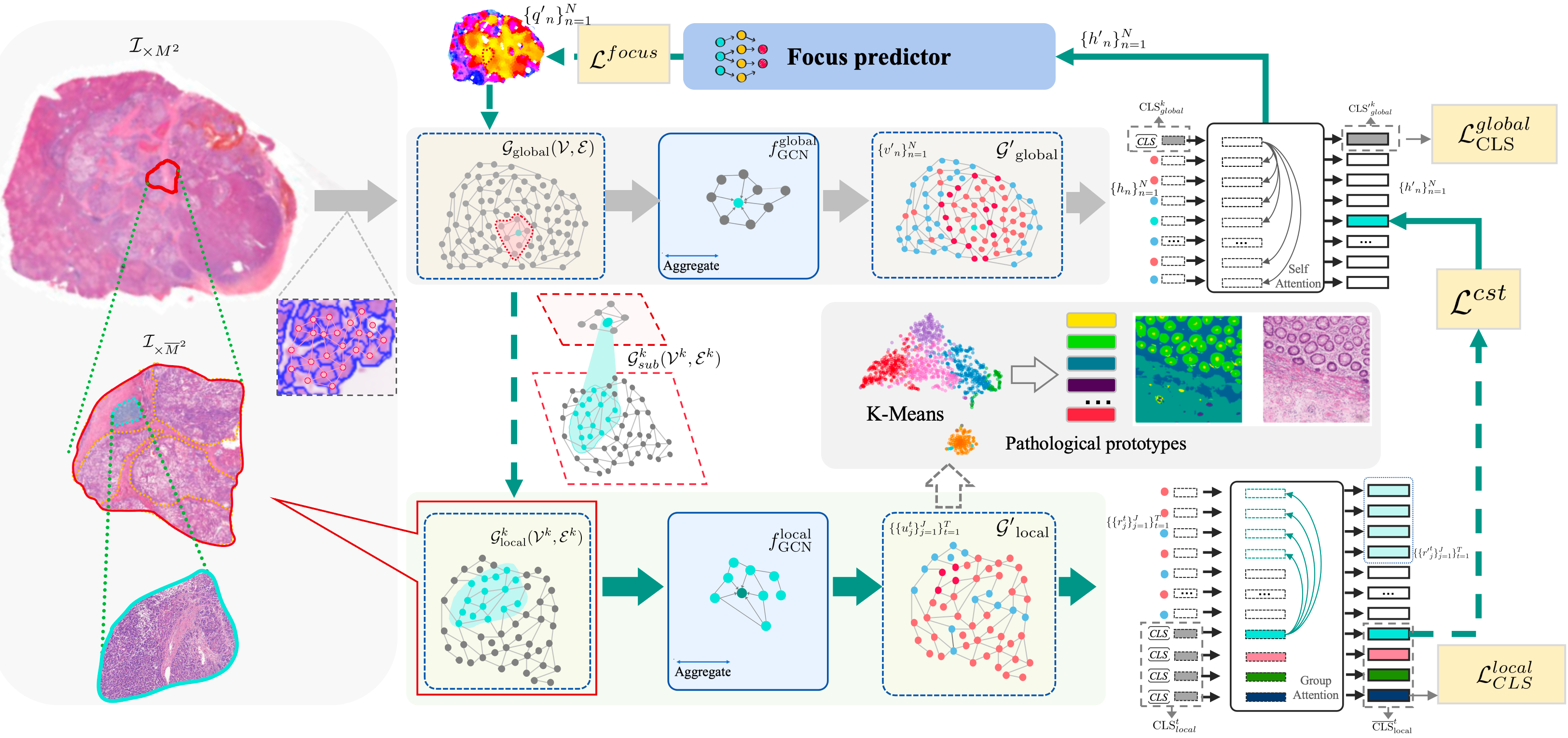}
    \caption{
    The SEW framework comprises a global branch, a focus predictor, and a detailed extraction branch. The global branch initially classifies the pathological thumbnail using loss function $\mathcal{L}^{global}_{\text{CLS}}$, while the focus predictor identifies relevant regions for classification based on the global branch's last layer features, guided by $\mathcal{L}^{focus}$. The detailed extraction branch then evaluates whether the magnified regions correspond to the lesion area using $\mathcal{L}^{local}_{\text{CLS}}$. Additionally, the feature similarity constraint $\mathcal{L}^{cst}$ between the global token and its corresponding local class token enhances the global branch's ability to extract discriminative features. 
    With the aggregated graph features, pathological prototypes are clustered to reinforce feature consistency across diverse WSI samples, a crucial step for tumor marker discovery.
    } 
    \label{fig:framework}
\end{figure*}


\subsection{Graph-based WSI Analysis}
Chen et al.~\cite{patchgcn} constructed a graph using features extracted from equally sized patches of WSI and applied Graph Convolutional Networks (GCNs) to learn structural features for survival prediction.
Similarly, Lu et al.~\cite{lu2022slidegraph+} utilized GCNs to predict HER2 status and breast cancer prognosis.
Lee et al.~\cite{teagraph} harnessed a Graph Attention Network to capture contextual features from a heterogeneous tumor environment.
Zhao et al.~\cite{fsgcnmilzhao2020predicting} utilized GCNs to learn bag-level representations for WSI analysis. 
To bolster the model’s global capabilities, Tang et al.~\cite{transgnn2024} introduced TransGNN, merging local structure with global long-range cross-attention for the prognosis prediction of hepatocellular carcinoma.
In contrast to the methods described above, we employ superpixels as graph nodes, preserving the original boundaries of distinct tissues. Furthermore, a multi-layer GCN framework is devised to capture features across various scales.

\subsection{Pathological Tumor Marker Mining}

Pathological markers offer invaluable insights into tumor diagnosis, prognosis, treatment response, and personalized care. Traditionally, pathologists identify novel tumor markers by adhering to established principles and guidelines, a process that is both labor-intensive and time-consuming, often proving costly~\cite{sharma2009tumor,das2023biomarkers}.  
Ye et al.~\cite{ye2023automated} annotated fine-grained tissue categories and trained a U-Net model to segment diverse tissue types, thereby assisting pathologists in identifying tumor markers.
Liang et al.~\cite{liver_biomarker_deepl} introduced a human-centric deep learning framework that utilizes CNNs to classify tissue patches, enabling pathologists to compare the differences between samples with good and poor prognoses.
Wagner et al.~\cite{crc_bio_marker_wagner2023transformer} developed a transformer-based pipeline for end-to-end biomarker prediction from pathology slides, leveraging the transformer's attention mechanism to facilitate biomarker mining.  
Ahn et al.~\cite{ahn2024histopathologic} applied MIL for prognosis prediction, subsequently clustering high-probability patches. Pathologists can then uncover tumor markers from these clustered features.  
However, the candidate features identified by these methods remain overwhelmingly numerous, making it difficult for pathologists to efficiently and swiftly pinpoint tumor markers from such an extensive pool. Furthermore, some deep learning-based approaches depend heavily on large-scale tissue annotations, which compromises their generalization ability.


\section{
Method
}
\fzl{
Pathology images are characterized by their large size and rich microscopic detail. To reduce interference from irrelevant features and enhance model inference speed, we propose a self-calibration enhanced framework for whole slide pathology image analysis, consisting of three components: a global branch, a focus predictor, and a detailed branch (see Fig.~\ref{fig:framework}). The global branch classifies using the pathological thumbnail, while the focus predictor identifies relevant regions based on the final layer features of the global branch. The detailed extraction branch then evaluates whether the magnified regions correspond to the lesion area. Lastly, a feature consistency constraint between the global and detailed branches ensures that the global branch focuses on relevant regions and extracts more discriminative features for identification.
}

\subsection{Global Superpixel Graph Classification}
\fzl{In the pathological diagnosis process, pathologists begin by identifying potential lesion areas at the thumbnail level. Inspired by this practice, we utilize a whole slide image (WSI) downsampled by a factor of $M^2$  as input in the global branch. Following superpixel segmentation to establish a graph, features are initially aggregated using graph neural networks. The transformer with global self-attention is then employed to extract global lesion structural features. These global features are directly used for WSI classification, effectively accelerating inference speed.}

\subsubsection{Superpixel Graph Building}

\fzl{
Given a thumbnail $\mathcal{I}_{\times M^2}$ of a pathology image \( \mathcal{I} \), the classic superpixel generation technique SLIC~\cite{achanta2010slic} is adopted to obtain the superpixel blocks.
}
\fzl{
Based on the previous superpixel block, we construct a superpixel graph as follows:
\begin{equation}
    \mathcal{G}_{\text{global}}(\mathcal{V},\mathcal{E}),
    \mathcal{V} = \{v_n\}_{n=1}^{N},
    \mathcal{E} = \{e_{n,n'}\}_{n=1}^{N}, \nonumber
\end{equation}
where $ v_n $ represents the $ n $-th node, $ e_{n,n'} $ denotes the edge between node $ v_n $ and its adjacent node $ v_{n'} $, $ N $ is the total number of nodes. For every node $v_n$, the color histogram of each color channel is extracted as its original color feature $z_n=[z^R_n,z^G_n,z^B_n]$. This operation is useful for feature aggregation and helps to reduce interference from massive similar features. Additionally, spatial information is incorporated into each node $v_n$ by concatenating the average position $p_n$ of the pixels in the $n$-th superpixel block with the color feature $z_n$. Therefore, the feature for each node $v_n$ is denoted as a composite feature $[z_n, p_n]$.
}

\subsubsection{Global Graph Classification}

We employ a graph convolutional network $ f_{\text{GCN}}^{\text{global}}() $ to perform convolution operations on the superpixel graph $ \mathcal{G}_{\text{global}}(\mathcal{V}, \mathcal{E}) $, resulting in the updated graph $\mathcal{G}'_{\text{global}}(\mathcal{V}', \mathcal{E}')=f_{\text{GCN}}^{\text{global}}(\mathcal{G}_{\text{global}}(\mathcal{V}, \mathcal{E}) )$ with aggregated features of adjacent nodes. It reduces feature diversity and the complexity of subsequent identification tasks. 

Then a transformer with a cross-attention module is adopted to model the global relations of the aggregated features $ \{v'_n\}_{n=0}^{N} $, without additional positional encoding due to the positional information already imposed by the GCN and the inherent translational invariance of self-attention.
The token embeddings $[ h_1, h_2, ..., h_N]$ for above transformer are obtained as follows:
\begin{align}
    [ h_1, h_2, ..., h_N] &= \text{Norm}(W^{\text{proj}}_{global}[v'_1, v'_2, ..., v'_N]), \nonumber \\ 
    W^{\text{proj}}_{global} & \in \mathbb{R}^{d \times d}, v'_n \in \mathbb{R}^{1 \times d}, \nonumber
\end{align}
where $W^{\text{proj}}_{global}$ is a learnable mapping matrix, $d$ is the dimension of the vector $v'_n$,  and $\text{Norm}()$ denotes normalize the $N\times d$ matrix into to $[0,1]$ according to the $d$-dimension.

Next, the global cross-attention transformer $f^{\text{global}}_{\text{Atten}}$ is adopted to model the global relationship among token embeddings $[ h_1, h_2, ..., h_N]$ along with an extra global class token $ \text{CLS}_{global}$ that reads out embeddings of all tokens as follows:
\begin{equation}
    [h'_1, ..., h'_N, \text{CLS}'_{\text{global}}] =f^{\text{global}}_{\text{Atten}}([h_1,..., h_N, \text{CLS}_{\text{global}}]), \nonumber
\end{equation}
where $[h'_1, ..., h'_N, \text{CLS}'_{\text{global}}]$ denotes the updated token embeddings. 

Then, an MLP classifier $f^{\text{global}}_{\text{MLP}}()$ is employed to perform initial classification at the thumbnail level, using the Cross-Entropy loss function as supervision:
\begin{equation}
    \mathcal{L}_{cls}^{global} = CE(y'_{cls},y_{gt}), y'_{cls}=f^{\text{global}}_{\text{MLP}}(\text{CLS}'_{\text{global}}), \nonumber
\end{equation}
where $y'_{cls}$ and $y_{gt}$ predicted the probability and ground truth of the WSI thumbnail, respectively.

\subsection{Focus Area Prediction}

Similar to the diagnostic process, pathologists first examine the overall lesion situation before focusing on critical areas for further scrutiny. In our approach, we propose a focus predictor to identify critical lesion areas for detailed feature extraction and identification.

To achieve this, we utilize a focus predictor $f^{\text{focus}}_{\text{MLP}}$() that takes the last layer token features $[h'_1, h'_2, ..., h'_N]$ of the global branch as input and predicts candidate areas. Consequently, we obtain the probability $q'_n$ of lesion areas $v_n$ as follows:
\begin{equation}
    q'_n = f_{\text{MLP}}^{\text{focus}}([h'_1, h'_2, ...,h'_N]),\nonumber
\end{equation}
The training of the focus predictor involves minimizing the Kullback-Leibler divergence between the predicted heatmap $Q'_{focus}$ and the ground truth $Q_{gt}$ as follows:
\begin{equation}
       \mathcal{L}^{focus} = D_{KL}(Q'_{focus}||Q_{gt}), Q'_{focus}=\{q'_n \}^N_{n=1}. \nonumber
\end{equation}
The focus predictor faces a cold start problem during early-stage training. To address this, the heatmap obtained by the Grad-CAM~\cite{2017Grad} from the global branch is adopted as pseudo label. Once the focus predictor and local branch demonstrate basic identification abilities, we use the prediction result of the local branch as a pseudo label. The local branch is trained using a detailed lesion area mask, enabling the extraction of more intricate tissue and cellular features. Using the local branch's output as a pseudo label for the focus predictor enhances focus on critical features.

\subsubsection{Top-K Sub-Graph Selection} 

Based on the predicted focus heatmap $Q'_{focus}=\{q'_n\}^N_{n=1}$, Top-K sub-graphs $\{\mathcal{G}_{\text{sub}}^k(\mathcal{V}^k,\mathcal{E}^k)\}_{k=1}^K$, with the highest average probability are selected. Each sub-graph corresponds to a group of nodes $\{v_{n'}\}^{N'}_{n'=1}$, as shown in Fig.~\ref{fig:framework}. The average probability is computed based on the predicted probability $q'_{n'}$ of all nodes in $\{v_{n'}\}^{N'}_{n'=1}$. It should be noted that the selected sub-graphs do not overlap with each other.

\subsection{Local Focus Area Calibration} 

In this section, the local branch is employed to extract features from the selected lesion areas that correspond to the Top-K sub-graphs. For each sub-graph $\mathcal{G}_{\text{sub}}^k(\mathcal{V}^k,\mathcal{E}^k)$, we isolate the corresponding amplified lesion region from the pathology $\mathcal{I}_{\times \overline{M}^2}$, which is downsampled by a factor of $\overline{M}^2$ from the original WSI $\mathcal{I}$. The extracted lesion region contains more detailed tissue and cellular features.

\subsubsection{Local Superpixel Graph Building} 
Similar to the global branch, the SLIC technique~\cite{achanta2010slic} is adopted to generate superpixel blocks for the corresponding area of each sub-graph $\mathcal{G}_{\text{sub}}^k(\mathcal{V}^k,\mathcal{E}^k)$. Additionally, the adjacent edge building technique is also adopted to construct the corresponding local graph $\mathcal{G}^k_{\text{local}}$. Using the local graph convolutional network $ f_{\text{GCN}}^{\text{local}}() $, we obtain the feature aggregation graph $\mathcal{G'}^k_{\text{local}}=f_{\text{GCN}}^{\text{local}}(\mathcal{G}^k_{\text{local}})$. It should be noted that the local branch does not require the addition of any position embedding.

The local graph $\mathcal{G}^k_{\text{local}}$ corresponds to the global sub-graph $\mathcal{G}^k_{\text{sub}}$ composed of $M$ nodes. Consequently, the nodes in $\mathcal{G}^k_{\text{local}}$ can be partitioned into $T$ groups, $\{\{u^t_j\}^J_{j=1}\}^T_{t=1}$, where $J$ denotes the number of nodes in each group. Remarkably, each group of nodes $\{u^t_j\}^J_{j=1}$ corresponds to a node in the global graph $\mathcal{G}_{\text{global}}$.

\subsubsection{Local Superpixel Graph Classification}

\fzl{
For each group of nodes $\{u^t_j\}^J_{j=1}$,  a mapping layer combined with a normalization operation is adopted to obtain the new embedding as follows:
\begin{align}
    [ r^t_1, r^t_2, ..., r^t_J] &= \text{Norm}(W^{\text{proj}}_{local}[u^t_1, u^t_2, ..., u^t_J]), \nonumber \\ 
    W^{\text{proj}}_{local} & \in \mathbb{R}^{J \times J}, u^t_j \in \mathbb{R}^{1 \times d}, \nonumber
\end{align}
where $W^{\text{proj}}_{local}$ is a learnable mapping matrix, $d$ is the dimension of the vector $u^t_j$,  and \text{Norm}() denotes normalize the $J\times d$ matrix into to $[0,1]$ according to the $d$-dimension.
}

\subsubsection{Group-wise Local Graph Classification}
The selected lesion area may contain multiple lesion types, such as different tumor types. Therefore, intra-group cross-attention is facilitated by adding $T$ $\{\text{CLS}^t_{\text{local}}\}^T_{t=1}$. Notably, the node representation $r^t_j$ in the $t$-th group is only cross-attentive with representations in the same group and the corresponding $\text{CLS}^t_{\text{local}}$. Simultaneously, interactions among the groups are calculated with the class tokens. The overall attention is computed as follows:
\begin{align}
    \text{CLS}'^t_{\text{local}} &= \text{GroupAtten}([ r^t_1, r^t_2, ..., r^t_J], \text{CLS}^t_{\text{local}}), \nonumber \\
    \overline{\text{CLS}}^t_{\text{local}} &= \text{Atten}(\{\text{CLS}'^t_{\text{local}}\}^T_{t=1},\text{CLS}'^t_{\text{local}}), \nonumber
\end{align}
where $\text{GroupAtten}()$ and $\text{Atten}()$ denote intra-group and inter-group cross-attention for class tokens, respectively.

Next, a local MLP classifier is adopted to classify the class token $\overline{\text{CLS}}^t_{\text{local}}$, which is given by:
\begin{equation}
\mathcal{L}^{local}_{\text{CLS}} =\frac{1}{T} \sum_{t=1}^T \text{CE}(\overline{y}^t,y^t_{gt}), \overline{y}^t=f^{\text{local}}_{\text{MLP}}( \overline{\text{CLS}}^t_{\text{local}}), \nonumber
\end{equation}
where $\overline{y}^t$ and $y^t_{gt}$ denote the predicted and ground truth categories for the $t$-th group, respectively. When supplied with a lesion area mask, $y^t_{gt}$ indicates whether the corresponding area belongs to the lesion area. On the other hand, if the WSI is labeled with areas of different types, $y^t_{gt}$ will be a multi-dimensional one-hot vector, denoting the type of tumor that corresponds to that area.

\subsubsection{Global and Local Consistency Constraint}
The group of nodes $\{u^t_j\}^J_{j=1}$ corresponds to a node in the global graph $\mathcal{G}_{\text{global}}$, as explained previously. Because the local graph $\mathcal{G}^k_{\text{local}}$ corresponds to the global sub-graph $\mathcal{G}^k_{\text{sub}}$, the $k$-th group corresponds to the $n$-th node in the global graph $\mathcal{G}$. Hence, we incorporate global and local consistency constraints to ameliorate the node feature extraction capabilities of $v_n$, which is formulated as follows:
\begin{equation}
\mathcal{L}^{\text{cst}} = D_{KL}(W^{\text{proj}}_{cls} \overline{\text{CLS}}^t_{\text{local}} || h'_n), \nonumber
\end{equation}
where $W^{\text{proj}}_{cls}$ represents a learnable matrix, $h'_n$ signifies the feature of node $v_n$ following a cross-attention operation.

\begin{table*}[t]
    \renewcommand\arraystretch{1.4}
    \footnotesize
	\centering
	\resizebox{\linewidth}{!}{
        \begin{tabular}{c|c|cccccccc|c}
        \toprule
        \multirow{1}{*}{Dataset} & \multicolumn{1}{c}{} &\multirow{1}{*}{Tea-Graph}  &\multirow{1}{*}{Patch-GCN} &\multirow{1}{*}{TransGNN} & \multirow{1}{*}{NIC} & \multirow{1}{*}{CLAM} & \multirow{1}{*}{HIPT} & \multirow{1}{*}{QuadTree} & \multirow{1}{*}{ZoomMIL} & \multirow{1}{*}{SEW} \\
        \midrule
        \multirow{2}{*}{CAMELYON16} & \emph{Acc.}  & $\underline{85.62\scriptstyle{\pm1.14}}$ & $80.06\scriptstyle{\pm0.81}$ & $82.83\scriptstyle{\pm0.81}$ & $80.83\scriptstyle{\pm0.94}$ & $83.16\scriptstyle{\pm0.86}$ & $\underline{85.57\scriptstyle{\pm1.05}}$ & $84.60\scriptstyle{\pm0.94}$& $84.42\scriptstyle{\pm1.33}$ & $\textbf{85.69}\scriptstyle{ \pm\textbf{0.85}}$ \\

         & \multicolumn{1}{c|}{\emph{Time}} & $692.32$ & $2118.26$ & $113.98$ & $2118.26$ & $113.98$ & $335.74$ & $\underline{71.35}$ & $428.19$ & $\emph{5.44}$ \\
        \cdashline{1-10}[2pt/2pt]

        \multirow{2}{*}{PANDA} & \emph{Acc.} & $\underline{82.17\scriptstyle{\pm 0.96}}$ & $80.76\scriptstyle{\pm2.06}$ & $79.62\scriptstyle{\pm1.16}$ & $78.19\scriptstyle{\pm2.71}$ & $80.80\scriptstyle{\pm1.07}$ & $80.69\scriptstyle{\pm3.23}$ & $76.60\scriptstyle{\pm1.96}$ & $\underline{81.38\scriptstyle{\pm1.28}}$ & $\textbf{82.62}\scriptstyle{\pm\textbf{1.97}}$  \\
        
         & \emph{Time} & $10.79$ & $51.86$ & $3.71$ & $51.86$ & $3.17$ & $8.93$ & $\underline{2.95}$ & $7.58$ & $\emph{1.85}$  \\
        \cdashline{1-10}[2pt/2pt]
        \multirow{2}{*}{BRCA} & \emph{Acc.} & $\underline{86.53\scriptstyle{\pm1.05}}$ & $85.15\scriptstyle{\pm1.77}$ & $85.06\scriptstyle{\pm1.30}$ & $84.02\scriptstyle{\pm1.80}$ & $85.82\scriptstyle{\pm{0.93}}$ & $\underline{87.26\scriptstyle{\pm2.25}}$ & $84.37\scriptstyle{\pm0.71}$ & $86.10\scriptstyle{\pm0.95}$ & $\textbf{87.44}\scriptstyle{\pm\textbf{0.94}}$ \\

         & \emph{Time} & $701.42$ & $968.46$ & $115.43$ & $1968.46$ & $115.43$ & $362.50$ & $\underline{80.50}$ & $505.20$ & $\emph{9.92}$ \\
        \cdashline{1-10}[2pt/2pt]

        \multirow{2}{*}{LUAD} & \emph{Acc.} & $\underline{81.12\scriptstyle{\pm1.54}}$ & $76.58\scriptstyle{\pm2.15}$ & $79.94\scriptstyle{\pm1.48}$ & $79.17\scriptstyle{\pm1.72}$ & $78.99\scriptstyle{\pm1.34}$ & $\underline{80.46\scriptstyle{\pm1.97}}$ & $76.93\scriptstyle{\pm1.88}$ & $78.68\scriptstyle{\pm1.18}$ & $\textbf{81.43}\scriptstyle{\pm\textbf{1.21}}$  \\
        
         & \emph{Time} & $578.63$ & $217.31$ & $99.03$ & $1217.31$ & $99.03$ & $179.95$ & $\underline{50.76}$ & $316.56$ & $\emph{8.06}$ \\
        \cdashline{1-10}[2pt/2pt]

       \multirow{2}{*}{HCC} & \emph{Acc.} & $86.23\scriptstyle{\pm1.40}$ & $81.09\scriptstyle{\pm2.37}$ & $85.11\scriptstyle{\pm1.85}$ & $86.65\scriptstyle{\pm2.02}$ & $\underline{87.83\scriptstyle{\pm1.53}}$ & $87.03\scriptstyle{\pm2.17}$ & $86.25\scriptstyle{\pm2.01}$ & $\underline{87.59\scriptstyle{\pm 1.99}}$ & $\textbf{87.93}\scriptstyle{\pm\textbf{0.63}}$ \\
        
         & \emph{Time} & $335.19$ & $504.32$ & $257.15$ & $3504.32$ & $257.15$ & $584.42$ & $\underline{101.54}$ & $757.79$ & $\emph{10.91}$ \\
     \cdashline{1-10}[2pt/2pt]
        
        \multirow{2}{*}{CRC} & \emph{Acc.} & $\underline{84.12\scriptstyle{\pm1.26}}$ & $81.09\scriptstyle{\pm3.04}$ & $83.10\scriptstyle{\pm1.44}$ & $81.95\scriptstyle{\pm1.34}$ & $82.57\scriptstyle{\pm1.66}$ & $\underline{83.73\scriptstyle{\pm2.17}}$ & $79.68\scriptstyle{\pm2.01}$ & $82.15\scriptstyle{\pm 1.99}$ & $\textbf{84.79}\scriptstyle{\pm\textbf{0.89}}$ \\
        
         & \emph{Time} & $564.33$ & $724.65$ & $437.13$ & $3801.75$ & $377.89$ & $644.10$ & $\underline{198.43}$ & $757.79$ & $\emph{9.82}$ \\
          \cdashline{1-10}[2pt/2pt]

          \multirow{2}{*}{GC} & \emph{Acc.} &$\underline{81.76\scriptstyle{\pm1.56}}$ & $77.59\scriptstyle{\pm1.78}$ & $80.69\scriptstyle{\pm0.95}$ & $77.29\scriptstyle{\pm1.68}$ & $80.44\scriptstyle{\pm1.05}$ & $81.83\scriptstyle{\pm1.59}$ & $79.22\scriptstyle{\pm1.88}$ & $\underline{82.07\scriptstyle{\pm 1.09}}$ & $\textbf{82.57}\scriptstyle{\pm\textbf{1.14}}$ \\
        
         & \emph{Time} & $261.28$ & $419.64$ & $215.84$ & $1664.90$ & $197.45$ & $297.76$ & $\underline{75.79}$ & $416.98$ & $\emph{5.47}$ \\

        \bottomrule
        \end{tabular}
    }

    \caption{
    \fzl{
    Performance comparison of different methods on various grading and prognostic datasets. The average slide-level accuracy (\%) and time (s) for each dataset are presented. The inference time for each slide includes pre-processing and prediction. The results with the best and second/third best results are marked in \textbf{bold} and \underline{underlined}, respectively. The inference time with the least/second least amount of time are represented in \textit{italics} and \underline{underlined}, respectively.
    }
    }
    \label{table:acc}
\end{table*}

\subsection{Pathological Prototype Vocabulary}
To ensure that tissues with the same semantics across diverse WSIs exhibit similar features, the pathological prototypes are derived in the following section. These prototypes can then be utilized to enhance classification performance and visualize the spatial distribution of tissue in WSIs. 

\subsubsection{Prototype Generation} 
After aggregating the nodes from all local graphs $\{\mathcal{G}_{\text{local}}^k\}^K_{k=1}$, node representations $\{u_i\}^{K \times T \times J}_{i=1}$ were clustered into $C$ clusters using the K-means clustering algorithm. These cluster center representations, denoted as $\{O_{c}\}^C_{c=1}$, form the pathological prototype vocabulary for the current WSI. The pathological prototype vocabulary can be utilized to reconstruct the lesion areas, providing a diagnostic reference for practical application. 

\subsection{Final Prediction and Model Optimization} 
To improve the WSI's final classification accuracy, we compute the fused multi-granularity feature vector, denoted as $H_{all}$. It includes global features, $K$ local sub-graph features, and cluster center features, combined as follows: 
\begin{align}
H_{all} &= W^{\text{mapping}}_{global} \text{CLS}_{global} \nonumber \\
& +W^{\text{mapping}}_{local} \frac{1}{K} \sum^K_{k=1} \text{CLS}^k_{local} +W^{\text{mapping}}_{proto} \frac{1}{C} \sum^C_{c=1} O_{c}, \nonumber
\end{align}
where $W^{\text{mapping}}_{global}$, $W^{\text{mapping}}_{local}$ and $W^{\text{mapping}}_{proto} $ are mapping matrices. With $H_{all}$, the final prediction $y'_{final}= f^{\text{all}}_{\text{MLP}}(H_{all})$ is obtained. 
The final classifier $f^{\text{all}}_{\text{MLP}}$ is trained with the Cross-Entropy loss function $\mathcal{L}^{all}=CE(y'_{final},y_{gt})$.

In summary, we optimize the focus predictor with $\mathcal{L}^{focus}$, the local branch with $\mathcal{L}^{local}_{cls}$, and the global branch with  $\mathcal{L}^{global}_{cls}$,  $\mathcal{L}^{cst}$ and $\mathcal{L}^{all}$ combined.

\subsection{Application on Tumor Marker Mining}

Pathological markers serve as objective indicators, facilitating early tumor diagnosis and intervention, thereby enhancing patient prognosis and reducing costs.  
The proposed SEW not only enables rapid and accurate classification by learning critical and discriminative features for pathological grading and prognosis tasks, but also facilitates tumor marker mining from two perspectives: feature cluster uniqueness and tissue spatial distribution.  

The focus predictor filters out most irrelevant feature areas, preserving only the most relevant regions for prognosis results. By further analyzing the differences in corresponding features of focused areas for favorable and unfavorable prognosis samples, the unique features of unfavorable samples reveal the tumor marker. In this paper, we employ the K-Means clustering algorithm to visualize different feature clusters. As illustrated in Fig.~\ref{fig:visualization} (a-c), the unique cluster corresponds to a novel tumor marker.

The detailed extraction branch, in conjunction with pathological prototypes, ensures that tissues with the same semantics across diverse WSIs are assigned to the same type. 
Consequently, the tissue spatial distribution differences between favorable and unfavorable prognosis samples can be identified. These differences in tissue spatial distribution offer valuable insights into the tumor marker from the perspective of tissue arrangement.

\section{Experiments}

\subsection{Experiment Setting}
\textbf{Datasets}. To evaluate the performance of the proposed method, we conducted tests across various types of cancer, assessing both the classification accuracy and speed for tasks such as grading and prognosis. Pathological data sets used in our experiments include PANDA\cite{pandadataset}, CAMELYON16\cite{camelyondataset}, BRCA\cite{lingle2016cancer}, and LUAD\cite{albertina2016cancer}. In addition to this, we have also collected three additional pathological datasets:
\textbf{HCC}: A dataset comprising 117 HE-stained pathological sections of Hepatocellular Carcinoma Cancer (HCC), labeled with five grades of lesion severity.
\textbf{GC}: A dataset comprising 123 HE-stained pathological sections of Gastric Cancer (GC), labeled with five grades of lesion severity.
\textbf{CRC}: A dataset comprising 343 HE-stained pathological sections of Colorectal Cancer (CRC), labeled with two prognostic categories.
Each section is annotated by professional pathologists with the lesion area and grade, as well as actual prognostic feedback.

\textbf{Implementation Details}. The devised model consists of two branches. We utilized SLIC for superpixel segmentation with a node count \( n = 1024 \), a GCN with 3 layers, and an output feature dimension \( d = 512 \). Layer normalization is applied between each layer, and residual connections are used. The transformer encoder is set to a depth of 12 layers, employing 4 attention heads. The batch size is set to 4, with the number of focused regions \( K = 4 \). Consequently, the effective batch size for the local branch is actually 16 due to the focus on \( K \) regions in the global branch. \fzl{
Stochastic Gradient Descent~\cite{ruder2016overview} is adopted for model training,
}with a momentum of 0.9 and a weight decay of \( 5 \times 10^{-4} \). The initial learning rate is set to 0.002 for the first two layers and 0.01 for the last layer. The training and inference of all methods run on a single RTX3090.

\begin{figure*}
    \includegraphics[width=\textwidth]{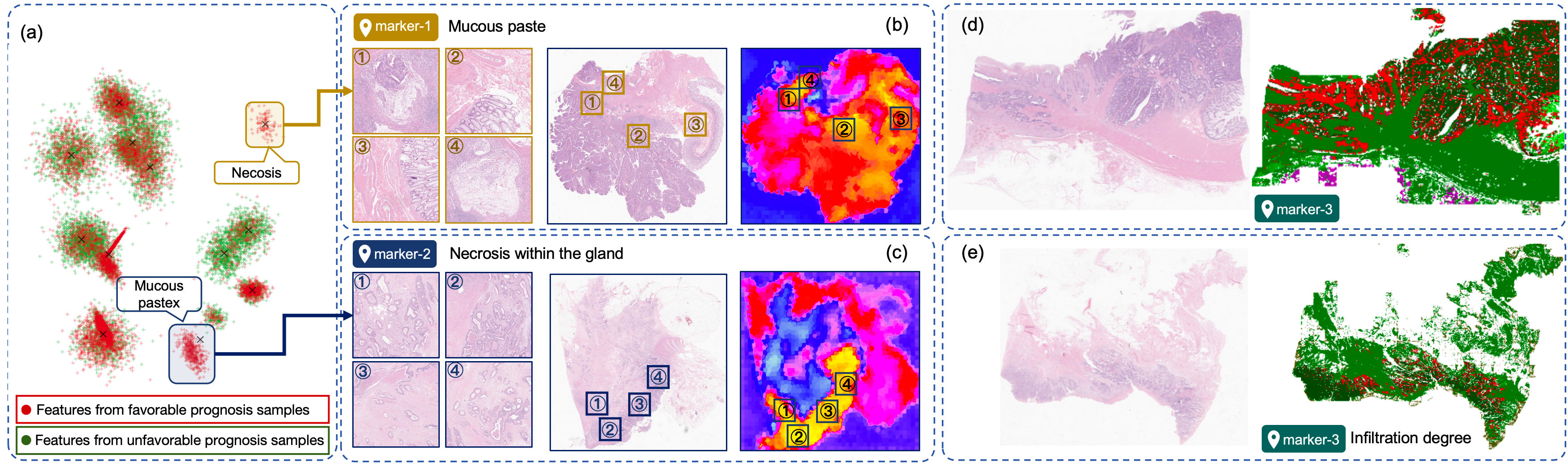}
    \caption{
Visualization of mined tumor markers in colorectal cancer samples:  
a) The SEW model is employed to extract pathological tissue-scale features from focused areas of colorectal cancer samples and perform clustering analysis, with particular emphasis on two feature clusters (with only \textcolor{red}{red} points) linked to poor prognosis.  
b) and c) showcase two novel tumor markers (verified by the pathologist) identified in the WSIs, along with their corresponding locations.  
d) presents the reconstructed WSI with pathological prototypes, where the spatial distribution of cancerous tissue (denoted in \textcolor{red}{red}) reveals the third tumor marker: the degree of tumor infiltration.} 
    \label{fig:visualization}
\end{figure*}

\subsection{
\fzl{Comparison with SOTA}
}
In this section, we conducted experiments using SEW on various datasets from different types of cancer and compared it with existing methods, including TeaGraph~\cite{teagraph}, TransGNN~\cite{transgnn2024}, Patch-GCN~\cite{patchgcn}, NIC~\cite{tellez2019neural}, CLAM~\cite{clamlu2021data}, HIPT~\cite{hipt2022scaling}, QuadTree~\cite{tangquadtree} and Zoom-MIL~\cite{zoommilthandiackal2022differentiable}. The experimental results are presented in the Table.~\ref{table:acc}.

SEW achieves the highest accuracy and the fastest speed on all types of cancer pathology image datasets. HIPT and Tea-Graph methods followed closely, achieving the next best results on different datasets, mainly due to HIPT's hierarchical structure that extracts sufficient features, and Tea-Graph's use of graph structures to fully explore the contextual features of pathology images. However, in terms of inference time, HIPT requires hundreds of seconds for the inference of each slide. Graph-based methods take a considerable amount of time due to the need to construct and calculate the graph across the entire slide. QuadTree and MIL-based methods accelerate by selecting partial regions within the WSI, but the acceleration is still limited. TransMIL and ZoomMIL cleverly pre-process to compress features in advance, reducing their inference time to just a few seconds, but the pre-processing takes hundreds of seconds. SEW constructs a graph at the thumbnail level and only zooms in on areas of interest, achieving fast and accurate predictions. Not only does SEW achieve the most outstanding performance on all types of slides, but it also requires only $5.44$-$10.97$ seconds for the inference of each WSI, which is an average improvement of $104.33$ times compared to the most advanced methods (the PANDA dataset, which is smaller in size, is temporarily not considered).

\subsection{Tumor Maker Mining and Visualization}

\textbf{Distinction feature clusters for tumor maker mining.}
The superior performance of SEW is primarily attributed to the self-calibrated focus predictor, which accurately identifies key regions of WSIs. In the prognosis task for colorectal cancer (CRC) patients, these key regions contain features strongly correlated with prognosis outcomes. 
We selected $100$ CRC patient cases with complete follow-up information, 50 of which had good prognoses, with no recurrence within five years, and 50 with poor prognoses, resulting in death within two years. Using the SEW model, which was well trained on  CRC dataset, we analyzed these patients' tissue slices, collected focused tissue-level features extracted from local subgraphs of all samples, and performed clustering on these features. 
\fzl{
 Fig. \ref{fig:visualization}(a) displays the clustering results, where red and green points represent features derived from poor and good prognosis samples, respectively. Distinct clusters (with only red color points) are observed for features from poor prognoses. Verified by the pathologist, these unique clusters correspond to mucinous lakes (\textbf{marker 1}) and necrosis within glands (\textbf{marker 2}), which are novel tumor markers for colorectal cancer.
}


\textbf{Tissue spatial distribution for tumor maker mining.}
Furthermore, the pathological prototypes can be utilized to reconstruct the entire WSI.
Fig.~\ref{fig:visualization}(d, e) illustrates the reconstruction results for good and poor prognosis samples. A notable difference is observed in the distribution of cancerous tissue (denoted in red) between the two types of samples.
The cancerous tissue invades and spreads into surrounding tissues, revealing the the degree of tumor infiltration (\textbf{marker 3}). This affirms the feasibility and effectiveness of using SEW for mining tissue distribution markers.
More visualization samples of mined tumor markers are given in the \emph{supplements}.

\subsection{Generalization Performance validation }
In this section, we designed a series of experiments to verify SEW's generalization performance. SEW was trained on the HCC dataset, and its model parameters were used as pre-trained ones. Then, it was fine-tuned on BRCA, LUAD, CRC, and GC datasets to test performance.

We compared the generalization performance of Patch-GCN, TransGNN, and Zoom-MIL with SEW. Through fine-tuning with the minimum number of epochs, SEW achieved an average accuracy improvement of $3.91\%$ over other methods. Moreover, the difference between a well-trained SEW and a fine-tuned SEW is only $0.07\sim0.64\%$. These results show that SEW's extracted features have strong generalizability for pathological image analysis.
 
\begin{table}[t]
    \footnotesize
	\centering
        \begin{tabular}{c|c|cccc}
        \toprule
        \multirow{1}{*}{Method} & \multicolumn{1}{c}{}  &\multirow{1}{*}{BRCA} &\multirow{1}{*}{LUAD} & \multirow{1}{*}{CRC} & \multirow{1}{*}{GC}  \\
        \midrule
        \multirow{2}{*}{Patch-GCN} & \emph{Acc.}  & ${82.96}$ & $74.78$ & $79.88$ & $75.45$  \\
         & \multicolumn{1}{c|}{\emph{Epoch}} & $19$ & $24$ & $16$ & $34$ \\
        \cdashline{1-6}[2pt/2pt]

        \multirow{2}{*}{TransGNN} & \emph{Acc.}  & $84.12$ & $78.01$ & $82.91$ & $79.44$  \\
         & \multicolumn{1}{c|}{\emph{Epoch}} & $17$ & $20$ & $10$ & $28$ \\
        \cdashline{1-6}[2pt/2pt]

        \multirow{2}{*}{ZoomMIL} & \emph{Acc.}  & $84.28$ & $75.26$ & $80.44$ & $80.15$  \\
         & \multicolumn{1}{c|}{\emph{Epoch}} & $18$ & $22$ & $14$ & $31$ \\
        \cdashline{1-6}[2pt/2pt]

        \multirow{2}{*}{SEW} & \emph{Acc.}  & $\textbf{87.07}$ & $\textbf{80.79}$ & $\textbf{84.49}$ & $\textbf{82.50}$  \\
         & \multicolumn{1}{c|}{\emph{Epoch}} & $12$ & $16$ & $7$ & $25$ \\
         
        \bottomrule
        \end{tabular}
    \caption{
    Generalization evaluation on various datasets with pretrained parameters on HCC. The number of fine-tuning epochs to converge and the corresponding accuracy are given to compare the generalization of various models.}
    \label{table:generalization}
\end{table}

\subsection{Ablation Study}

\textbf{Superpixel vs Patch.}
We take the superpixel blocks generated by the superpixel method and utilize them as nodes in the graph. In contrast to the approach of employing patches as graph nodes in TeaGraph or PatchGNN, we removed the superpixel segmentation module from SEW and directly used 16x16 patches as graph nodes to test the performance on the HCC dataset. As presented in Table ~\ref{table:preprocess}, superpixel nodes exhibit better boundary adhesion and bring about a significant improvement.
\begin{table}[t]
\centering
\footnotesize 
\begin{tabular}{llccc}
\toprule
Method & Dataset &  Acc.(\%)           & Time (s)  & AUC\\

\midrule
Patch       & HCC       & 83.39         & 7.95      & 0.82  \\
            & CAMELYON16& 83.70         & 4.93    & 0.82  \\
Superpixel  & HCC       &\textbf{87.93} & 10.91     & 0.88  \\
            & CAMELYON16& \textbf{85.69}& 5.44      & 0.87  \\
\bottomrule
\end{tabular}
\caption{
Comparison between superpixel blocks and patches for graph construction on the HCC and Camelyon16.
}
\label{table:preprocess}
\end{table}

\begin{figure}[!t]
    \centering
    \includegraphics[width=0.47\textwidth]{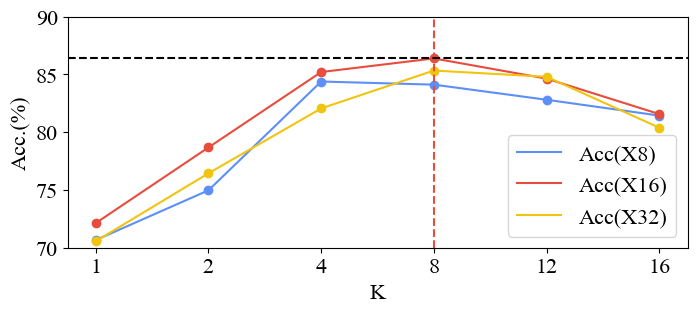} 
    \caption{
    \fzl{
     The accuracy curve for various magnification rates (8x, 16x, and 32x) with different numbers of focus areas.
    }
    }
    \label{fig:k}
\end{figure}

\textbf{
Focusing areas number K and  magnification.} SEW can achieve high accuracy with only a small number of focusing areas for magnification. To explore the impact of the number of K on the performance of SEW, we verified the performance of K taking 1, 2, 4, 8, 12 and 16 on the HCC dataset. As shown in Fig.~\ref{fig:k}, when K is between 4 and 8, the accuracy of SEW no longer significantly improves with the increase of K, indicating that SEW has obtained enough detailed information for reliable pathological diagnosis when K=8. When K=16, the performance declined, which may be due to the introduction of redundant information that interferes with the information of key areas and affects the final diagnostic performance. At the same time, we also studied the impact of the magnification of local branches on performance. In the local branches, we used 8, 16 and 32 times magnified information to enhance the global branch, and on the HCC dataset, the 16 times magnification method achieved better performance due to better alignment with the information of the global branch.

\textbf{
\fzl{Different components impact.}
} The global branch can sense the information of the entire slice at the thumbnail level and use the local branch's detailed information for self-enhancement. We compared the performance of the global branch alone and the enhanced one. In the forward process, we compared using only the Grad-CAM score with using the focus predictor for focusing. In the backward process, we removed the $\mathcal{L}_{cst}^{ret}$ constraints to determine if the local branch's output enhanced the global branch's information extraction. In Table ~\ref{table:ablation-study}, using the local branch and the dual-branch framework's consistency constraint significantly improved the model's performance. The focusing effect of the focus predictor led to a substantial enhancement compared to using Grad-CAM alone. This improvement results from the local branch enhancing the global branch, which depends on the focus location's accuracy.

\begin{table}[!t]
\centering
\footnotesize 
\begin{tabular}{cccccccc} 
\toprule
  \multicolumn{2}{c}{Branch} & \multicolumn{2}{c}{Focus} & \multicolumn{1}{c}{Back} &  \multicolumn{3}{c}{Metrics}\\
 \cmidrule(lr){1-2} \cmidrule(lr){3-4} \cmidrule(lr){5-5} \cmidrule(lr){6-8}
    Glob. &Loc. & $grad$ & $q$ & $\mathcal{L}_{cst}$ & {Acc.(\%)} &{Time(s)} &{AUC}\\
\midrule
    \cmark &      &       &       &        & 75.21          & 3.77  & 0.76 \\
    \cmark &\cmark&\cmark &       &        & 79.64          & 9.64  & 0.82 \\
    \cmark &\cmark&       &\cmark &        & 83.22          & 9.55  & 0.87 \\
    \cmark &\cmark&       &\cmark &  \cmark& \textbf{86.39} & 10.91 & 0.88 \\
\bottomrule
\end{tabular}
\caption{Ablation study of components in SEW, `Glob.' denotes the global branch, `Loc.' denotes the local branch, `grad.' denotes the grad-cam, and `q' denotes the result of the focus predictor.}
\label{table:ablation-study}
\end{table}

\section{Conclusion}

In this study, we introduced SEW, a method that can extract features closely associated with pathological image analysis, achieving promising classification results based on these features. Moreover, SEW is capable of identifying biomarkers at both the tissue and tissue distribution levels. On a colorectal cancer dataset, SEW successfully discovered two novel tumor biomarkers, demonstrating the potential of artificial intelligence in exploring new prognostic tumor biomarkers.In the future, we will focus on developing more user-friendly methods and tools for biomarker mining to facilitate their clinical application.

\clearpage
{
    \small
    \bibliographystyle{ieeenat_fullname}
    \bibliography{main}
}
\end{document}